\documentclass{article}

\usepackage{PRIMEarxiv}

\usepackage[utf8]{inputenc} 
\usepackage[T1]{fontenc}    
\usepackage{hyperref}       
\usepackage{url}            
\usepackage{booktabs}       
\usepackage{amsfonts}       
\usepackage{nicefrac}       
\usepackage{microtype}      
\usepackage{lipsum}
\usepackage{fancyhdr}       
\usepackage{graphicx}       
\graphicspath{{media/}}     
\usepackage{hyperref}
\usepackage{url}
\usepackage{graphicx}
\usepackage{booktabs}       
\usepackage{amsfonts}       
\usepackage{nicefrac}       
\usepackage{microtype}      
\usepackage{xcolor}         
\usepackage{amsmath}
\usepackage{graphicx}
\usepackage{booktabs}
\usepackage[table]{xcolor}
\usepackage{amssymb}
\usepackage{xcolor}
\usepackage{float}
\usepackage{subcaption}
\usepackage{algorithm}
\usepackage{algorithmic}
\usepackage{amsthm}
\title{Kinematics-Induced Multimodal 3D Human Pose Estimation with Subject-Level Privacy}

\author{
  Kaushik Bhargav Sivangi \\
  Bio-AIm Lab \\
  School of Computing Science \\
  University of Glasgow\\
  2836578s@student.gla.ac.uk \\
   \And
  Fani Deligianni \\
  Bio-AIm Lab \\
  School of Computing Science \\
  University of Glasgow\\
  fani.deligianni@glasgow.ac.uk \\
}

\begin{document}
\maketitle

\begin{abstract}
Multimodal 3D Human Pose Estimation (3D HPE) combines complementary information from RGB, LiDAR, and mmWave radar, but models trained on  correlated observations from the same individuals, raise privacy risks overlooked by record level analysis. We present a unified framework for multimodal 3D HPE that couples kinematics-induced sensor fusion with subject level privacy auditing and private training. First, our multimodal model aligns modality specific joint representation, injects skeletal structure and adaptively aggregates complementary sensor evidence for accurate pose prediction. Second, we formulate a black-box subject membership inference attack for 3D HPE, complemented by an empirical pointwise maximal leakage analysis, which characterizes how individual attack score outcomes change inference about the membership outcome. Third, we instantiate user-level differential privacy via Action Temporal Stratification, a population weighted within-subject sampling strategy that enforces action and temporal coverage. We evaluate our framework on the MM-Fi dataset across three diverse experimental protocols. Source-code will be released upon acceptance.
\end{abstract}

\section{Introduction}

Three dimensional human pose estimation (3D HPE) is a central problem in computer vision, supporting applications in healthcare monitoring, rehabilitation, sports analysis, and human computer interaction. The majority of existing approaches operate on RGB video, benefiting from large scale image datasets and mature two stage pipelines that detect 2D keypoints before lifting them into 3D space \cite{chen2017_3DHPE,dang2023reconstructing,qi2019attentive, qi2020stc, qi2019sports, qi2020imitative, moon2019camera, wehrbein2021probabilistic, pavlakos2017coarse}, with more recent methods incorporating temporal convolutions, transformers, kinematic priors, and diffusion based refinement \cite{pavllo20193d, zheng20213d, zhang2022mixste, zhu2023motionbert, peng2024ktpformer, shan2023diffusion}. However, RGB model deployment remains limited in unconstrained settings such as hospitals and assisted living environments due to performance degradation under occlusions, poor illumination, viewpoint variations and privacy-concerns. \cite{dang2023reconstructing,yang2024privacy}.

Because no single modality is reliable across all conditions, multimodal systems combining RGB, LiDAR, and millimeter-wave radar have emerged to leverage their complementary strengths \cite{furst2021hperl, an2022mri, yang2023mm}. However, effectively integrating these heterogeneousness modalities remains challenging. It requires resolving the differing feature distributions and noise characteristics of RGB, LiDAR, and radar into a common semantic space, while retaining the complementary, modality-specific evidence each sensor provides \cite{chen2025x,2024Chae_FusionObjectDet,furst2021hperl,yang2023mm}.  


A second, underexplored, concern in this setting is privacy. Human pose data captured through RGB, LiDAR, or radar can be linked back to individual subjects, yet the computer vision community has repeatedly been shown to overlook or mischaracterize this risk \cite{chen2019understanding,baldassini2025dontcallit,2025zhang_pedestrianprivacy}. Differential privacy (DP) provides a principled and mathematically grounded framework \cite{dwork2014algorithmic} but applying it to visual prediction tasks via DP stochastic gradient descent (DP-SGD) incurs substantial utility loss \cite{golatkar2022mixed, zhou2021bypassing,sivangi2026dp2dpose,fang2023improved}. Moreover, standard record-level DP protects individual samples but not the repeated participation of a subject across many records. This is especially problematic for LiDAR and radar-based human motion, whose four-dimensional spatio-temporal sequences are strongly correlated across frames from the same subject, so record-level guarantees can substantially understate an individual's true privacy risk \cite{ghazi2021user,chua2024mind}.

Despite growing interest in both multimodal pose estimation and privacy preserving learning, the two problems have largely been studied in isolation. Audits of what a trained multimodal pose model reveals about its subjects remain scarce, with prior privacy risk analysis in this space typically framed as membership or attribute inference under record level assumptions \cite{gupta2021membership,sivangi2026dp2dpose}. This leaves open the question of how to design a multimodal pose estimation framework whose privacy is assessed and enforced at the level of the subject rather than the observation, and whether the resulting protection can be quantified beyond aggregate membership accuracy.

In this work, we present a unified framework that couples kinematics induced multimodal fusion with subject level privacy auditing and private training for 3D-HPE. Our contributions are threefold: 
\textbf{i)} First, we introduce a kinematics-induced fusion module featuring a \textbf{dual-track design with adaptive gating} that aligns modality-specific joint representations from RGB, LiDAR, and radar, and injects skeletal structure through fixed anatomical connectivity combined with learned joint dependencies. It adaptively aggregates the resulting evidence into a shared representation that outperforms each individual modality and achieves state-of-the-art performance. 
\textbf{ii)} Second, we formulate a black box subject level membership inference attack tailored to the regression nature of HPE and complement it with an empirical pointwise maximal leakage analysis \textbf{incorporating Dirichlet smoothing} \cite{saeidian2023pointwise}, which characterizes how individual attack outcomes shift an adversary's belief about subject membership, offering a finer grained view of leakage than threshold based attack metrics alone. 
\textbf{iii)} Third, we instantiate differential privacy at the subject rather than the sample level, coupled with \textbf{private quantile-based adaptive clipping}, following the general user level formulation of \cite{chua2024mind} and the correlated sampling perspective of \cite{ghazi2021user}. Furthermore, we introduce \textbf{Action Temporal Stratification}, a population weighted within subject sampling strategy that preserves an unbiased subject gradient while enforcing coverage across actions and time.

\section{Related Work}
\subsection{3D HPE using multimodal sensors}

RGB based 3D HPE has become a central research direction in computer vision, largely driven by the availability of large scale image and video datasets and well established 2D pose detectors. Early approaches typically adopted a two-stage pipeline: first detecting 2D keypoints, then lifting these 2D poses into 3D space\cite{qi2019attentive, qi2020stc, qi2019sports, qi2020imitative, moon2019camera, wehrbein2021probabilistic, pavlakos2017coarse}, with recent work using temporal convolutions, transformers, kinematic priors, and diffusion models to model spatial-temporal pose structure \cite{pavllo20193d, zheng20213d, zhang2022mixste, zhu2023motionbert, peng2024ktpformer, shan2023diffusion}. Despite strong benchmark performance, RGB-based methods remain sensitive to occlusion, illumination, resolution, viewpoint changes, and scene clutter because they do not directly observe 3D geometry.


LiDAR and mmWave Radar have been explored as complementary sensing modalities for 3D HPE. LiDAR provides explicit 3D geometry and illumination invariance, enabling long-range markerless motion capture and robust 3D HPE from point clouds \cite{li2022lidarcap, zhang2024lidarcapv2, an2025pre,zheng2022multi}. However, LiDAR point clouds of humans are often sparse, incomplete, and sensitive to distance, self-occlusion, and sensor placement. mmWave Radar offers additional advantages under poor lighting and partial occlusion, and has been used for skeletal pose estimation, mesh recovery, diffusion-based pose estimation, and multi-view sensing \cite{sengupta2020mm, xue2021mmmesh, fan2024diffusion, choi2025mvdoppler,zhu2025probRadarm3f}. Yet Radar point clouds remain extremely sparse, noisy, and anatomically coarse compared with RGB and LiDAR observations.


Multimodal 3D HPE aims to combine these complementary cues for improved robustness and accuracy. Prior work has explored RGB-LiDAR pose estimation and multimodal human sensing datasets such as mRI and MM-Fi, which provide synchronized sensor streams for evaluating individual modalities and fusion strategies \cite{furst2021hperl, an2022mri, yang2023mm}. However, most multimodal systems require multiple sensors during inference, increasing hardware cost, calibration complexity, synchronization overhead, and deployment constraints. This motivates teacher-student settings in which a multimodal teacher learns from complementary sensors during training, while a single-modality student performs efficient inference using only one available sensor.

\subsection{Privacy-preservation in computer vision}
Growing evidence indicates that privacy preservation within computer vision community has frequently been ignored or misunderstood \cite{2025zhang_pedestrianprivacy,baldassini2025dontcallit}. The field has largely relied on fragmented, improvised anonymization techniques lacking formal guarantees. Rigorous alternatives such as differential privacy (DP) offer mathematical bounds, yet applying standard DP stochastic gradient descent to visual prediction tasks incurs a substantial utility loss ~\cite{golatkar2022mixed}. Vision models suffer due to severe long-tail class imbalance that amplifies noise on minority classes, ie. rare poses, high dimensionality that reduces the signal-to-noise ratio of the gradient vectors \cite{fang2023improved,zhou2021bypassing}, and continuous pixel values that create non-smooth loss landscapes violating Lipschitz continuity~\cite{fang2023improved}. Consequently, DP-SGD performance relies critically on loss smoothness, dimensionality, and clipping calibration ~\cite{lebensold2024privacy,sivangi2026dp2dpose}.

Furthermore, it has been increasingly recognized that standard record level DP is insufficient when training data contains strong temporal correlations or multi sample contributions from individual subjects \cite{ghazi2021user,chua2024mind,2019_Kawamoto_DistributionPrivacy}. This concern is particularly acute for LiDAR and Radar based human motion signatures, which are inherently four dimensional spatio-temporal point cloud sequences, with each subject contributing frames exhibiting substantial temporal autocorrelation.
This paradigm requires user level, rather than sample level, privacy protection, using advanced mechanisms such as correlated sampling to handle structured per-user dependencies \cite{ghazi2021user}. 

\section{Methodology}
\subsection{Overall Architecture}

\begin{figure}
    \centering
    \includegraphics[width=\textwidth]{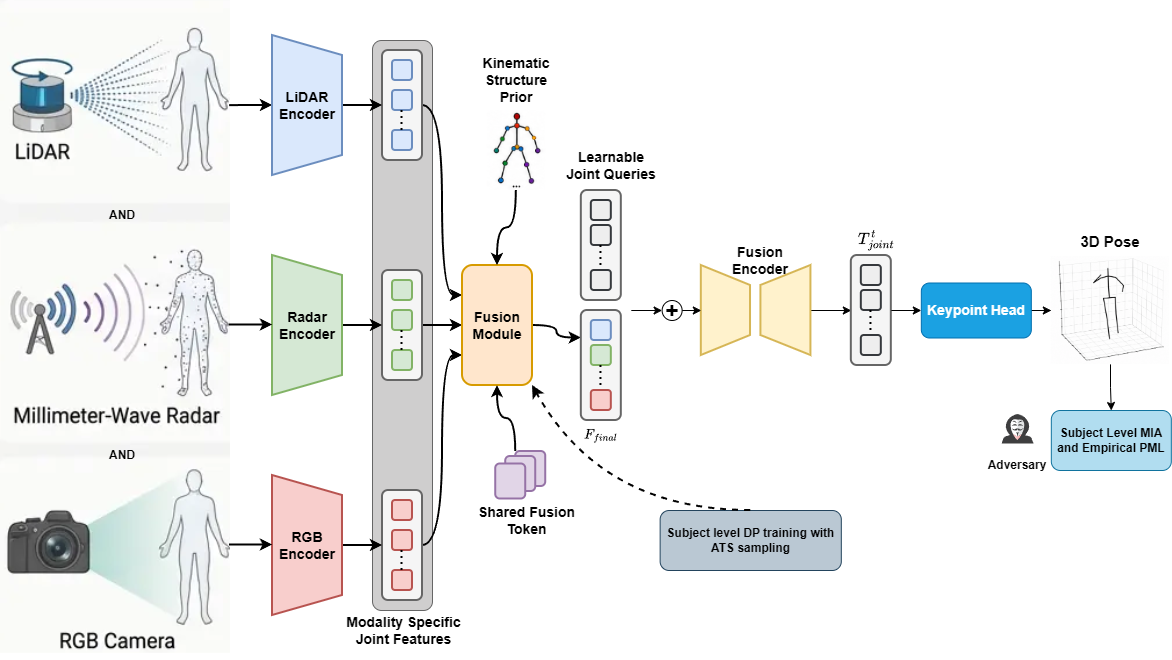}
    \caption{Overall framework of the proposed privacy aware multimodal 3D HPE method. RGB, LiDAR, and mmWave radar keypoint representations are integrated through the kinematics induced fusion module to predict the 3D pose. The model is trained on Action Temporal Stratified Subject level DP, while the final model is attacked through subject membership inference and empirical pointwise maximal leakage.}
    \label{fig:graph-structure}
\end{figure}

The proposed architecture employs a heterogeneous modality specific encoders followed by a kinematics induced multimodal fusion module for 3D HPE. We utilize heterogeneous backbones to extract modality-specific joint representations: a DSTFormer-based encoder for RGB derived 2D keypoint sequences, and PointTransformerV2-based encoders for LiDAR and Radar point clouds. The modality specific representations are subsequently integrated through our kinematics induced fusion module for multimodal 3D HPE. This module uniquely maintains a modality specific joint representations together with shared fusion tokens, refining modality specific features through skeletal priors while simultaneously updating the shared representation through adaptive gating. This hierarchical design allows modality contributions to vary across anatomical joints and input samples through joint wise adaptive fusion weights.
To preserve articulated structure, learnable joint queries are processed by a transformer-based keypoint head to produce one token for each of the $J=17$ body joints \cite{li2021tokenpose}. The multimodal model predicts the final 3D pose from these joint tokens using a regression head. The overall architecture is given in Figure \ref{fig:graph-structure}.

\subsection{Fusion Module}
To effectively integrate LiDAR, Radar, and RGB modalities, we design a kinematics-induced fusion module that operates in three levels: modality alignment, kinematic-token fusion, and joint-wise aggregation. This architecture preserves modality-specific characteristics while progressively extracting a shared representation consistent across sensors and aware of skeletal constraints.

\paragraph{Modality Alignment:} {The three modalities exhibit different feature distributions due to their sensing mechanisms. LiDAR captures geometric structure, Radar offers sparse but robust measurements in any environment, and RGB provides rich visual and motion cues. Fusing these heterogeneous features can lead to unstable optimization and poor cross-modal agreement. We address this through a lightweight alignment stage that performs cross-modal calibration. Let $F_L,F_R,F_I \in \mathbb{R}^{J \times D }$ denote the features extracted from LiDAR, Radar and RGB derived keypoint backbone where $J$ is the number of joints and $D$ is the feature dimension. We append a learnable modality embedding $e_m$ to each stream and apply multi-headed attention (MHA) jointly across the three modalities at each joint location which is given as:

\begin{equation}
\begin{split}
    \tilde{F_m} & = F_m + e_m,\quad m \in {L,R,I} \\
    [X_L, X_R, X_I] & = [F_L, F_R, F_I] + MHA ([\tilde{F_L}, \tilde{F_R}, \tilde{F_I}])
\end{split}
\end{equation}

This interaction encourages the three streams to align at identical anatomical locations, allowing each modality to incorporate complementary evidence before the primary fusion stage.
}

\paragraph{Kinematic Injection and Dual Track Design:} {Post alignment, we employ a dual track design that maintains  modality specific joint features alongside shared fusion tokens. The modality specific joint features retain local evidence from each sensor, while the shared tokens accumulate information that is jointly supported across all modalities. Within each stream, joint features are first refined via a kinematic propagation mechanism based on the human skeleton structure. 
Let $X_m \in \mathbb{R}^{J \times D}$ denote the modality specific joint features for modality $m$. Based on \cite{peng2024ktpformer}, we define a modality dependent affinity matrix by combining the fixed skeletal adjacency $A_{ske}$ with a learnable residual adjacency $\hat{A}_m$. This is then symmetrized to encode both known anatomical connectivity and learned long-range joint dependencies. We specifically update the kinematic refinement with the shared token which overall constitutes to calculating a gating coefficient. Formally, 

\begin{equation}
    \begin{split}
        A_m & = \frac{(A_{ske} + \hat{A}_m) + (A_{ske} + \hat{A}_m)^T}{2} \\
        & \\
        G_m & = \sigma(MLP([X_m;S])) \\
        H_m & = X_m + A_m(G_m \odot X_m)
    \end{split}
\end{equation}
where $S$ is the initialized shared token ($S \in \mathbb{R}^{K \times D}$), $G_m$ is an adaptive gating coefficient. This allows each modality to refine its joint representation in a structure-aware manner. Moreover, the gate is adjusted by current shared representation, implying the amount of kinematic refinement is not fixed and adapts according the cross-modal fusion state of the sample. 


The modality specific  features are then fused at each joint through adaptive weights. The model predicts per-joint fusion coefficients from the features themselves, allowing the contribution of each modality to vary across body parts and samples. This is given as:
\begin{equation}
    F = \sum_{m} \alpha_m \odot H_m; \quad \sum_m \alpha_m = 1
\end{equation}
where the fusion weights $\alpha_m$ are produced by a lightweight MLP followed by softmax across modalities. This enables the contribution of each modality to adapt across anatomical joints and input samples. 

The shared fusion tokens $S$ are updated by attending to the fused representation
\begin{equation}
    S' = S + MHA(S, F, F)
\end{equation}
The shared tokens act as a cross-modal information pool. Since it receives information from fused anatomical representation, it gradually accumulates the consensus level information. The shared tokens are then fed back to each modality stream through cross-attention given as:

\begin{equation}
\begin{split}
\Delta_m &= MHA(H_m, S', S') \\
    X_m^{'} &= X_m + \Psi([H_m, F, \Delta_m])
    \end{split}
\end{equation}
where $\Psi(\cdot)$ denotes MLP layer. This ensures each of the modality-specific feature are more informative and cross-modally aware. We finally obtain the modality specific joint features $X_m^{'}$, together with shared token $S'$. A final joint-wise weighted fusion is then applied to produce the final fused representation:
\begin{equation}
    F_{final} = \sum_{m \in {L,R,I}}  \alpha_m^{final} \odot X_m'
\end{equation}
This final representation summarizes the complementary strengths of the three modalities while preserving the anatomical consistency through kinematic refinement.
}

\subsection{Action Temporal Stratified Subject-Level Differential Privacy}
\label{sec:ats_dp}

Multi-modal human pose dataset contain many observations contributed by the same subject, including temporally overlapping windows within individual actions. A record level DP mechanism that treats each temporal window as an independent privacy unit protects the inclusion or removal of one window, but does not directly protect the inclusion or removal of a subject's complete contribution. Towards this, we instantiate subject level differential privacy where we treat each subject's entire records as the privacy unit following the general user wise formulation of \cite{chua2024mind} and prior user level private training with multiple samples per user \cite{ghazi2021user}. For our setting, all the multi-modal pose windows belonging to one subject constitute a single privacy unit.

\paragraph{Subject-level Neighborhood:}
Let the training dataset consist of $N$ subjects,
\begin{equation}
    \mathcal{D}
    =
    \biguplus_{s=1}^{N}\mathcal{D}_{s},
    \qquad
    \mathcal{D}_{s}
    =
    \{z_{s,1},\ldots,z_{s,M_s}\},
    \label{eq:subject_dataset}
\end{equation}

where $z_{s,i}$ denotes one multimodal pose window and $M_s$ is the number of available windows for subject $s$. Two datasets are neighbors if they differ by the complete contribution of one subject.
\begin{equation}
    \mathcal{D}' = \mathcal{D}\setminus\mathcal{D}_{s}.
    \label{eq:subject_adjacency}
\end{equation}

\paragraph{Differential Privacy:}
A randomized training mechanism $\mathcal{M}$ satisfies subject-level $(\epsilon,\delta)$-differential privacy if, for every neighboring $\mathcal{D},\mathcal{D}'$ and measurable output set $\mathcal{O}$,
\begin{equation}
    \Pr[\mathcal{M}(\mathcal{D})\in\mathcal{O}]
    \leq
    e^{\epsilon}
    \Pr[\mathcal{M}(\mathcal{D}')\in\mathcal{O}]
    +\delta .
    \label{eq:subject_dp}
\end{equation}
This is the standard differential privacy definition \cite{dwork2014algorithmic}, with neighborhood specialized to complete subject removal rather than removal of a single pose window.

\paragraph{Subject-wise private gradient:} At private optimization step $t$, each of the $N$ subject slots is included independently with probability $\rho$,
\begin{equation}
    I_{t,s}\sim\operatorname{Bernoulli}(\rho),
    \qquad
    \mathcal{S}_{t}
    =
    \{s:I_{t,s}=1\}.
    \label{eq:poisson_subject_sampling}
\end{equation}
This constitutes Poisson sampling at the protected subject level. For each sampled subject, $K$ pose windows are used to construct one subject-gradient estimate $g_{t,s}$ and clipping is applied only after this within subject aggregation \cite{chua2024mind}, where multiple records belonging to a sampled user are first averaged into one user contribution.

The resulting subject vector is clipped once with a global $\ell_2$ bound
$C_t$,
\begin{equation}
    \bar g_{t,s}
    =
    g_{t,s}
    \min\left(
        1,
        \frac{C_t}{\|g_{t,s}\|_2}
    \right),
    \qquad
    \|\bar g_{t,s}\|_2\leq C_t .
    \label{eq:subject_clipping}
\end{equation}
The privatized aggregate is
\begin{equation}
    \widetilde g_t
    =
    \frac{1}{\rho N}
    \left[
        \sum_{s\in\mathcal{S}_t}\bar g_{t,s}
        +
        \sigma_g C_t \xi_t
    \right],
    \qquad
    \xi_t\sim\mathcal{N}(0,I),
    \label{eq:private_gradient}
\end{equation}
where $\rho N$ is the public expected number of sampled subjects. Since subsequent optimizer operations depend only on privatized gradients, they are post-processing of the private mechanism \cite{dwork2014algorithmic}. To avoid manual tuning of this hyperparameter, we adapt the private quantile-based adaptive clipping mechanism from \cite{andrew2021differentially} to complete subject gradients, as detailed in Appendix ~\ref{sec:appendix}.


\paragraph{Why stratify windows within a subject?} Uniform within-subject sampling provides an unbiased estimate of a subject's mean gradient, but pose sequences exhibit substantial semantic and temporal redundancy. With a finite record budget $K$, a random draw may therefore
over-represent a frequently occurring action or a short temporal region while missing other behaviors of the same subject. Such heterogeneous sampling can increase stochastic-gradient variance, motivating stratified sampling in stochastic optimization \cite{zhao2014accelerating,lu2021variance}.

For fixed parameters $\theta$, let the complete pre-clipping gradient of
subject $s$ be
\begin{equation}
    \mu_s(\theta)
    =
    \frac{1}{M_s}
    \sum_{z\in\mathcal{D}_s}
    \nabla_\theta \ell(\theta;z).
    \label{eq:full_subject_gradient}
\end{equation}
Our goal is to approximate $\mu_s(\theta)$ using only $K \ll M_s$ windows without changing the intended complete subject mean gradient. We therefore partition $\mathcal{D}_s$ by action and temporal position and construct the population-weighted estimator
\begin{equation}
    g_s^{\mathrm{ATS}}
    =
    \sum_{h=1}^{H_s}
    \frac{M_{s,h}}{M_s}g_{s,h},
    \label{eq:ats_overview}
\end{equation}
where $M_{s,h}$ is the population of stratum $h$ and $g_{s,h}$ is its sampled mean gradient. As proven in Appendix ~\ref{sec:appendix} (Lemma 1), population weighting ensures an unbiased pre-clipping subject gradient in expectation, enforcing robust action and temporal coverage without altering the true subject mean.
In particular,
\begin{equation}
    \mathbb{E}\!\left[g_s^{\mathrm{ATS}}\mid\mathcal{D}_s,\theta\right]
    =
    \mu_s(\theta),
    \label{eq:ats_unbiased_main}
\end{equation}
Thus, ATS changes how the within subject sampling budget is allocated, rather than changing the intended pre clipping gradient in expectation. A formal derivation of this result is provided in Appendix \ref{sec:appendix}.

\paragraph{Action Temporal Stratification (ATS):} For subject $s$, we partition the legal windows into disjoint and exhaustive strata, 
\begin{equation}
    \mathcal{D}_s
    =
    \biguplus_{h=1}^{H_s}\mathcal{D}_{s,h},
    \qquad
    M_{s,h}=|\mathcal{D}_{s,h}|,
    \qquad
    \sum_{h=1}^{H_s}M_{s,h}=M_s .
    \label{eq:strata_partition}
\end{equation}
Strata are constructed using action identity and chronological position
within each action. In practice, we first allocate sampling slots across
non-empty actions and subsequently divide each action sequence into contiguous
temporal bins.

Let $A_s$ be the number of non-empty actions for subject $s$. We reserve one sampling slot for every action and distribute the remaining
$R_s=K-A_s$ slots approximately in proportion to proportion to the number of windows in each action. For an action $a$ containing $M_{s,a}$ windows, define
\begin{equation}
    q_{s,a}
    =
    R_s\frac{M_{s,a}}{M_s},
    \qquad
    n_{s,a}
    =
    1+\lfloor q_{s,a}\rfloor+\Delta_{s,a},
    \label{eq:action_allocation}
\end{equation}
where $\Delta_{s,a}\in\{0,1\}$ is assigned to actions with the largest fractional remainders such that $\sum_a n_{s,a}=K$. Windows within each action are sorted chronologically and divided into contiguous temporal bins according to allocated number of slots $n_{s,a}$.

More generally, suppose $n_{s,h}\geq1$ samples are drawn uniformly from stratum $h$, with
\begin{equation}
    \sum_{h=1}^{H_s}n_{s,h}=K.
\end{equation}
The mean gradient within the sampled stratum is
\begin{equation}
    g_{s,h}
    =
    \frac{1}{n_{s,h}}
    \sum_{j=1}^{n_{s,h}}
    \nabla_{\theta}
    \ell(\theta;Z_{s,h,j}),
    \label{eq:stratum_gradient}
\end{equation}
where $Z_{s,h,j}$ is sampled uniformly from $\mathcal{D}_{s,h}$. Our
Action Temporal Stratified estimator is
\begin{equation}
    g^{\mathrm{ATS}}_{s}
    =
    \sum_{h=1}^{H_s}
    \frac{M_{s,h}}{M_s}
    g_{s,h}.
    \label{eq:ats_gradient}
\end{equation}
Equivalently, every selected draw from stratum $h$ receives weight
\begin{equation}
    w_{s,h,j}
    =
    \frac{M_{s,h}}{M_s n_{s,h}}.
    \label{eq:ats_weight}
\end{equation}
Population weighting allows us to enforce action and temporal coverage without assigning disproportionate objective weight to smaller strata. Importantly, ATS modifies only the construction of the pre-clipping subject gradient. The resulting vector is still clipped once at the subject level as in Eq.\ref{eq:subject_clipping}. Consequently, its contribution remains bounded by $C_t$, while the outer Bernoulli subject sampling and Gaussian mechanism remain unchanged.
Appendix~\ref{sec:appendix_Lemma1} (Proposition 1) formally establishes that replacing uniform sampling with ATS does not alter the outer subject-level privacy accounting or require modifications to our Rényi differential privacy composition.
For fixed $\rho$, effective noise multiplier $\sigma$, total private steps $T$,$\delta$, and adaptive clipping configuration, ATS uses the same subject level privacy accounting as random within subject sampling. We provide the formal privacy argument in Appendix \ref{sec:appendix}. We adapt private adaptive subject level clipping following \cite{andrew2021differentially}; $\sigma_g$ in Eq.\ref{eq:private_gradient} denotes the resulting gradient-noise multiplier after jointly accounting for the private clipping-count query. We use Rényi differential privacy to compose the repeated sampled gaussian mechanisms \cite{mironov2017renyi}. The complete adaptive-clipping derivation, noise calibration are provided in Appendix~\ref{sec:appendix}. 

\subsection{Subject Level Membership Inference and Pointwise Maximal Leakage}

We evaluate membership evidence using the same subject-level units from private training. While membership inference measures whether an adversary can distinguish training from non training subjects, pointwise maximal leakage (PML) characterizes how individual score outcomes shift membership beliefs. 

\paragraph{Threat model and membership experiment:} Following the privacy game formulation of \cite{salem2023sok} and the subject membership setting from \cite{suri2022subject}, let $\mathcal{S}_{\mathrm{tr}}$ denote the subjects included in the main training pool and define indicator as 
\begin{equation}
    B_s=\mathbf{1}\{s\in\mathcal{S}_{\mathrm{tr}}\}.
    \label{eq:mia_membership}
\end{equation}
Membership only denotes inclusion in the training dataset. The attacker has prediction only access to the fixed final model $f_{\theta_T}$ and $Q$ query windows from the target subject, together with the 3D pose annotations. The adversary does not have access to model parameters, gradients, intermediate features, or metadata that would reveal any membership. This is a ground truth aware black-box threat model. Membership labels are defined externally by the experiment split and are never obtained from the trained checkpoint.

\paragraph{Subject level membership score:} As 3D HPE is a regression problem, we use prediction error as the membership inference signal\cite{yeom2018privacy,gupta2021membership}. For each query window $i$, we compute the MPJPE error $e_i$ to derive the subject-level membership inference score $S_Q$:
\begin{equation}
 \qquad
 S_Q(s)=-\frac{1}{Q}\sum_{i=1}^{Q}e_i
 \label{eq:mia_score}
\end{equation}
For each subject, the $Q$ queried windows produce a score where a higher value corresponds to a lower pose error and stronger membership evidence under this attack. We use subject level ROC-AUC as the primary threshold independent metric. For threshold-based evaluation, we define $\hat B_s=1\{S_Q(s)\geq\tau\}$, where $\tau$ is selected by stratified subject level cross fitting using balanced accuracy on remaining folds.

\paragraph{Empirical Pointwise Maximal Leakage:} While MIA measures how well the attack scores distinguish about the membership class, PML measures the increase in inference confidence of the adversary associated with that particular outcome. We define the audit game with a balanced membership prior as $P(B=0)=P(B=1)=1/2$ \cite{saeidian2023pointwise}. Since $S_Q$ is continuous, we use a small set of fixed score boundaries, and this is declared independently of the evaluated scores, to obtain the discrete evidence variable $Y\in\{0,\ldots,H-1\}$. Let $n_{b,h}$ be the number of subjects with membership label $b$ and let score lie in bin $h$, and $N_b=\sum_h n_{b,h}$. To prevent zero probability estimates in small subject setting, we use Dirichlet-1/2 smoothing:
\begin{equation}
    \hat p_b(h)=
    \frac{n_{b,h}+\frac{1}{2}}
         {N_b+\frac{H}{2}} .
    \label{eq:pml_prob}
\end{equation}

For the balanced binary membership bit, the PML\cite{saeidian2023pointwise} definition specializes as:
\begin{equation}
 \ell(B\!\rightarrow\!h)=
 \log\frac{2\max\{\hat p_0(h),\hat p_1(h)\}}
 {\hat p_0(h)+\hat p_1(h)}, \qquad
 0\leq\ell(B\!\rightarrow\!h)\leq\log 2 .
 \label{eq:pml}
\end{equation}

A value of zero denotes that the observed outcome does not change the balanced membership belief, whereas $log2$ corresponds to complete disclosure of the membership variable. As smoothing assigns positive probability even to bins unobserved in both classes, the primary empirical summary takes the maximum only over bins containing at least one observed subject.

\begin{equation}
 \widehat{\ell}^{\,obs}_{\max}
 =
 \max_{h:n_{0,h}+n_{1,h}>0}\ell(B\!\rightarrow\!h).
 \label{eq:pml_max}
\end{equation}
This is empirical estimate for the specified MIA and is not to be interpreted as a formal guarantee.

\section{Experimental Results}

\subsection{Results}

\subsubsection{Effect of Multimodal Fusion}

The main experimental setup is described in Appendix \ref{sec:appendixB}. As shown in Table \ref{tab:protocol3}, under protocol 3, which considers all activities, the proposed fusion model achieves the lowest MPJPE scores across all three evaluation splits. Compared with the strongest unimodal implementation in each split, fusion reduces MPJPE by 5.46mm, under random split, 10.81mm under cross environment split and 11.05mm under cross subject split. The larger improvements under cross environment and cross subject settings suggest that combining complementary information from RGB, LiDAR, and Radar is particularly beneficial under distribution shifts. Results for protocols 1 and 2 are provided in Appendix \ref{sec:prot12}, while additional modality ablations and comparisons with a BeV based fusion model are presented in Appendix \ref{sec:bimodal}, \ref{sec:Bev} respectively.

\begin{table}[tbp]
\centering
\renewcommand{\arraystretch}{1.08}
\footnotesize
\setlength{\tabcolsep}{2.8pt}

\resizebox{\linewidth}{!}{
\begin{tabular}{l l cc cc cc}
\toprule
\textbf{Model} &
\textbf{Modality} &
\multicolumn{2}{c}{\textbf{Random}} &
\multicolumn{2}{c}{\textbf{Cross-Environment}} &
\multicolumn{2}{c}{\textbf{Cross-Subject}} \\
\cmidrule(lr){3-4}
\cmidrule(lr){5-6}
\cmidrule(lr){7-8}
&
&
\textbf{MPJPE} $\downarrow$ &
\textbf{PA-MPJPE} $\downarrow$ &
\textbf{MPJPE} $\downarrow$ &
\textbf{PA-MPJPE} $\downarrow$ &
\textbf{MPJPE} $\downarrow$ &
\textbf{PA-MPJPE} $\downarrow$ \\
\midrule

\multicolumn{8}{c}{\textbf{Protocol 3: All Activities}} \\
\midrule
\multicolumn{8}{c}{\textit{Existing Methods}} \\
\midrule

MM-Fi
\cite{yang2023mm}
& RGB
& 279.0 & 81.2
& 288.6 & 84.1
& 285.3 & 81.9 \\

MM-Fi
\cite{yang2023mm}
& LiDAR
& 92.5 & 61.5
& 303.8 & 133.0
& 103.8 & 61.5 \\

MM-Fi
\cite{yang2023mm}
& Radar
& 117.0 & 57.3
& 161.6 & 73.7
& 129.7 & 60.0 \\

MM-Fi
\cite{yang2023mm}
& RGB+LiDAR
& 87.5 & 59.8
& 192.1 & 89.6
& 104.6 & 62.1 \\

X-Fi
\cite{chen2025x}
& RGB
& 93.9 & 60.3
& -- & --
& -- & -- \\

X-Fi
\cite{chen2025x}
& LiDAR
& 167.1 & 103.2
& -- & --
& -- & -- \\

X-Fi
\cite{chen2025x}
& Radar
& 127.4 & 69.8
& -- & --
& -- & -- \\

X-Fi
\cite{chen2025x}
& RGB+LiDAR
& 93.0 & 59.7
& -- & --
& -- & -- \\

X-Fi
\cite{chen2025x}
& RGB+Radar
& 88.8 & 57.3
& -- & --
& -- & -- \\

X-Fi
\cite{chen2025x}
& LiDAR+Radar
& 109.8 & 63.4
& -- & --
& -- & -- \\

X-Fi
\cite{chen2025x}
& RGB+LiDAR+Radar
& 88.4 & 57.2
& -- & --
& -- & -- \\

\midrule
\rowcolor{blue!8}
\multicolumn{8}{c}{\textit{Our Implementations}} \\
\midrule

\rowcolor{blue!4}
PointTransformerV2
& LiDAR
& 46.24 & 36.13
& 60.70 & 40.12
& 55.28 & 38.21 \\

\rowcolor{blue!4}
PointTransformerV2
& Radar
& 61.09 & 45.27
& 78.42 & 58.58
& 61.80 & 47.11 \\

\rowcolor{blue!4}
DSTFormer
& RGB
& 44.34 & 32.84
& 63.30 & 48.03
& 48.70 & 36.87 \\

\rowcolor{green!6}
Fusion
& RGB+LiDAR+Radar
& \textbf{38.88} & 29.45
& \textbf{49.89} & 40.03
& \textbf{37.65} & 26.04 \\

\bottomrule
\end{tabular}
}

\caption{Comparison on Protocol 3 using all activities. Lower MPJPE and
PA-MPJPE indicate better performance.}
\label{tab:protocol3}
\end{table}

\subsubsection{ATS Subject Level DP and Attack inference}
Table \ref{tab:dp_mia_pml_results} demonstrates the subject-level privacy-utility tradeoff. Stronger DP suppresses membership evidence: $\epsilon = 1$ yields near chance ROC-AUC and zero empirical PML at a sever utility cost. As $\epsilon$ increases pose accuracy improves alongside attack distinguishability. The non-private model ($\epsilon = \infty$) achieves the best MPJPE but also the strongest membership signal, with 7 out of the 8 held out subjects falling in the higher error half of the attack scores. With only 8 subjects held-out, only the non-private model and $\epsilon = 200$ are individually distinguishable from chance. PML changes more coarsely than ROC-AUC as it relies on discretized score outcomes. 
\begin{table}[tbp]
\centering
\renewcommand{\arraystretch}{1.12}
\footnotesize
\setlength{\tabcolsep}{6pt}

\begin{tabular}{c c c c}
\toprule
\textbf{$\epsilon$} &
\textbf{MPJPE (mm)} $\downarrow$ &
\textbf{ROC-AUC} &
\textbf{Empirical PML (nats)} \\
\midrule

1   & 1081.32 & 0.48 & 0.0 \\
16  & 314.85 &  0.59&0.14  \\
64  & 278.58 &  0.64& 0.14 \\
100 &269.86  &  0.65& 0.29\\
200 & 247.72 &  0.73& 0.29 \\
$\infty$ & 37.65 & 0.83& 0.44\\
\bottomrule
\end{tabular}

\caption{
Privacy--utility and empirical leakage results under the MM-Fi Protocol 3 cross-subject setting. Each row corresponds to a subject-level
differentially private model trained at the indicated privacy budget $\epsilon$ and non-dp baseline is also provided. MPJPE measures 3D pose estimation utility, ROC-AUC measures subject-membership discrimination, and empirical PML characterizes leakage in the corresponding membership-evidence channel. Lower MPJPE and PML are better, while ROC-AUC closer to $0.5$ indicates weaker membership distinguishability.
}
\label{tab:dp_mia_pml_results}
\end{table}

\section{Conclusion}
In this work we present a framework for multimodal 3D human pose estimation that combines kinematics-induced fusion with subject-level DP training, and empirical membership inference. Overall, our fusion model improves the pose accuracy over individual modalities. The proposed Action Temporal Stratification constructs population-weighted subject gradients, while membership inference and PML provide complementary evidence of model privacy. The results show utility degradation under subject-level DP, particularly at stricter privacy budgets. Our evaluation is limited to the predefined MM-Fi split of subjects with eight non-member subjects. Future work should investigate more effective private adaptation techniques and evaluate on broader subject cohorts.

\bibliographystyle{unsrt}  
\bibliography{references}  

\section{Appendix}
\label{sec:appendixB}
\subsection{Experimental Setup}
\label{sec:experimentalsetup}
\paragraph{Dataset and Implementation:} {We utilize MM-Fi \cite{yang2023mm}, a comprehensive multimodal dataset designed for non-intrusive 4D human sensing. The dataset comprises over 320K synchronized frames across five modalities: RGB, depth, LiDAR, mmWave Radar, and WiFi CSI over 40 subjects. It encompasses 27 action categories, consisting of 14 daily activities and 13 rehabilitation activities. We specifically use LiDAR, Radar and RGB modalities for our experiments. MM-Fi is particularly challenging for Radar human pose estimation due to its sparse Radar point clouds containing fewer than 100 points per frame compared to other benchmarks. To test the robustness of our approach, we adopt the dataset’s three standard splits: random split, cross-subject split, and cross-environment split across all three protocols. We implement all our models on two NVIDIA RTX 4090 GPUs. We use PointTransformerV2 \cite{wu2022point} as backbone for LiDAR and Radar point clouds while DSTFormer \cite{zhu2023motionbert} serves as backbone for RGB based model. Additionally, in the RGB setting we follow the two stage pipeline of generating 3D pose from 2D estimates. MM-Fi dataset consists of 2D keypoints in prior extracted from HRNet-48 backbone and we leverage this. We use Adam optimizer for our experiments with a batch size of 32. The total epochs are set to 50. The learning rate is set to 1e-3 which is then multiplied by 0.1 after 40 epochs. We report Mean Per Joint Position Error (MPJPE) and Procrustes-aligned MPJPE (PA-MPJPE), following the standard 3D HPE evaluation protocols.

Private models are evaluated only on MM-Fi protocol 3 under cross subject split with 32 training subjects treated as members and 8 held out subjects treated as non members. We train for $T=2000$ private steps with subject sampling probability $\rho = 0.25, \delta=10^{-5}, and K=64$ ATS selected windows per sampled subject.

\subsection{Additional results of fusion method on Protocol 1 and 2}
\label{sec:prot12}
\begin{table}[H]
\centering
\renewcommand{\arraystretch}{1.08}
\footnotesize
\setlength{\tabcolsep}{2.8pt}

\resizebox{\linewidth}{!}{
\begin{tabular}{l l cc cc cc}
\toprule
\textbf{Model} &
\textbf{Modality} &
\multicolumn{2}{c}{\textbf{Random}} &
\multicolumn{2}{c}{\textbf{Cross-Environment}} &
\multicolumn{2}{c}{\textbf{Cross-Subject}} \\
\cmidrule(lr){3-4}
\cmidrule(lr){5-6}
\cmidrule(lr){7-8}
&
&
\textbf{MPJPE} $\downarrow$ &
\textbf{PA-MPJPE} $\downarrow$ &
\textbf{MPJPE} $\downarrow$ &
\textbf{PA-MPJPE} $\downarrow$ &
\textbf{MPJPE} $\downarrow$ &
\textbf{PA-MPJPE} $\downarrow$ \\
\midrule

\multicolumn{8}{c}{\textbf{Protocol 1: Daily Activities}} \\
\midrule
\multicolumn{8}{c}{\textit{Existing Methods}} \\
\midrule

Point Transformer
\cite{zhao2021point,fan2024diffusion}
& Radar
& 73.09 & 55.60
& 88.28 & 68.79
& 75.96 & 58.70 \\

DiffPose
\cite{gong2023diffpose,fan2024diffusion}
& Radar
& 73.44 & 56.83
& 86.35 & 66.87
& 70.31 & 54.12 \\

mmDiff
\cite{fan2024diffusion}
& Radar
& 65.26 & 50.35
& 82.73 & 63.87
& 65.62 & 50.23 \\

MM-Fi
\cite{yang2023mm}
& Radar
& 109.8 & 55.6
& 166.2 & 73.9
& 128.4 & 58.7 \\

MM-Fi
\cite{yang2023mm}
& RGB
& 263.3 & 80.0
& 276.6 & 83.0
& 267.7 & 81.2 \\

MM-Fi
\cite{yang2023mm}
& LiDAR
& 98.1 & 65.2
& 192.3 & 100.4
& 110.1 & 66.2 \\

MM-Fi
\cite{yang2023mm}
& RGB+LiDAR
& 94.3 & 63.9
& 159.6 & 84.5
& 109.4 & 65.8 \\

Balanced Multimodal
\cite{qi2025towards}
& Radar
& 102.89 & 52.21
& -- & --
& -- & -- \\

Balanced Multimodal
\cite{qi2025towards}
& RGB
& 63.61 & 35.75
& -- & --
& -- & -- \\

Balanced Multimodal
\cite{qi2025towards}
& LiDAR
& 66.95 & 45.70
& -- & --
& -- & -- \\

Balanced Multimodal
\cite{qi2025towards}
& RGB+LiDAR
& 52.93 & 34.96
& -- & --
& -- & -- \\

Balanced Multimodal
\cite{qi2025towards}
& RGB+Radar
& 63.04 & 35.74
& -- & --
& -- & -- \\

Balanced Multimodal
\cite{qi2025towards}
& LiDAR+Radar
& 66.85 & 45.11
& -- & --
& -- & -- \\

\midrule
\rowcolor{blue!8}
\multicolumn{8}{c}{\textit{Our Implementations}} \\
\midrule

\rowcolor{blue!4}
PointTransformerV2
& LiDAR
& 47.60 & 38.73
& 66.68 & 47.68
& 52.47 & 39.02 \\

\rowcolor{blue!4}
PointTransformerV2
& Radar
& 61.65 & 49.42
& 83.27 & 60.55
& 62.14 & 46.12 \\

\rowcolor{blue!4}
DSTFormer
& RGB
& 54.10 & 41.84
& 63.38 & 47.04
& 56.06 & 43.59 \\

\rowcolor{green!6}
Fusion
& RGB+LiDAR+Radar
& \textbf{40.60} & \textbf{32.91}
& \textbf{55.62} & \textbf{42.53}
& \textbf{45.38} & \textbf{34.98} \\

\midrule
\multicolumn{8}{c}{\textbf{Protocol 2: Rehabilitation Activities}} \\
\midrule
\multicolumn{8}{c}{\textit{Existing Methods}} \\
\midrule

MM-Fi
\cite{yang2023mm}
& RGB
& 291.6 & 83.6
& 301.5 & 85.3
& 304.3 & 82.6 \\

MM-Fi
\cite{yang2023mm}
& LiDAR
& 94.9 & 60.4
& 186.0 & 103.5
& 103.9 & 60.3 \\

MM-Fi
\cite{yang2023mm}
& Radar
& 124.3 & 55.3
& 168.0 & 73.0
& 142.2 & 57.4 \\

MM-Fi
\cite{yang2023mm}
& RGB+LiDAR
& 90.1 & 58.9
& 153.5 & 87.1
& 103.5 & 60.2 \\

\midrule
\rowcolor{blue!8}
\multicolumn{8}{c}{\textit{Our Implementations}} \\
\midrule

\rowcolor{blue!4}
PointTransformerV2
& LiDAR
& 42.76 & 32.97
& 61.01 & 46.43
& 41.79 & 32.43 \\

\rowcolor{blue!4}
PointTransformerV2
& Radar
& 59.76 & 44.93
& 81.04 & 63.07
& 58.66 & 43.94 \\

\rowcolor{blue!4}
DSTFormer
& RGB
& 41.59 & 32.70
& 54.93 & 41.38
& 42.51 & 33.02 \\

\rowcolor{green!6}
Fusion
& RGB+LiDAR+Radar
& \textbf{38.64} & 30.67
& \textbf{53.00} & 39.64
& \textbf{41.39} & 29.97 \\

\bottomrule
\end{tabular}
}

\caption{Comparison on Protocol 1 (Daily Activities) and Protocol 2
(Rehabilitation Activities). Lower MPJPE and PA-MPJPE indicate better
performance.}
\label{tab:protocol12}
\end{table}

\subsection{Bimodal Ablations}
\label{sec:bimodal}
\begin{table}[H]
\centering
\renewcommand{\arraystretch}{1.08}
\footnotesize
\setlength{\tabcolsep}{4pt}

\resizebox{\linewidth}{!}{
\begin{tabular}{l l l c c}
\toprule
\textbf{Protocol} &
\textbf{Split} &
\textbf{Bimodal Fusion} &
\textbf{MPJPE} $\downarrow$ &
\textbf{PA-MPJPE} $\downarrow$ \\
\midrule

\multicolumn{5}{c}{\textbf{Protocol 1: Daily Activities}} \\
\midrule

P1 & Random & RGB+LiDAR & 42.24 & 34.02 \\
P1 & Random & RGB+Radar & 46.17 & 35.69 \\
P1 & Random & LiDAR+Radar & 44.89 & 34.02 \\

P1 & Cross-Environment & RGB+LiDAR & 55.68 & 42.68 \\
P1 & Cross-Environment & RGB+Radar & 60.91 & 44.73 \\
P1 & Cross-Environment & LiDAR+Radar & 58.66 & 45.04 \\

P1 & Cross-Subject & RGB+LiDAR & 47.69 & 36.21 \\
P1 & Cross-Subject & RGB+Radar & 48.60 & 36.44 \\
P1 & Cross-Subject & LiDAR+Radar & 49.28 & 36.93 \\

\midrule
\multicolumn{5}{c}{\textbf{Protocol 2: Rehabilitation Activities}} \\
\midrule

P2 & Random & RGB+LiDAR & 41.66 & 31.96 \\
P2 & Random & RGB+Radar & 42.80 & 32.68 \\
P2 & Random & LiDAR+Radar & 42.11 & 33.41 \\

P2 & Cross-Environment & RGB+LiDAR & 52.15 & 39.83 \\
P2 & Cross-Environment & RGB+Radar & 52.39 & 39.88 \\
P2 & Cross-Environment & LiDAR+Radar & 57.62 & 43.84 \\

P2 & Cross-Subject & RGB+LiDAR & 42.73 & 32.26 \\
P2 & Cross-Subject & RGB+Radar & 43.33 & 33.13 \\
P2 & Cross-Subject & LiDAR+Radar & 43.80 & 34.01 \\

\midrule
\multicolumn{5}{c}{\textbf{Protocol 3: All Activities}} \\
\midrule

P3 & Random & RGB+LiDAR & 41.96 & 33.50 \\
P3 & Random & RGB+Radar & 44.29 & 34.16 \\
P3 & Random & LiDAR+Radar & 43.21 & 34.07 \\

P3 & Cross-Environment & RGB+LiDAR & 54.47 & 41.37 \\
P3 & Cross-Environment & RGB+Radar & 55.86 & 42.42 \\
P3 & Cross-Environment & LiDAR+Radar & 67.03 & 49.71 \\

P3 & Cross-Subject & RGB+LiDAR & 45.89 & 34.72 \\
P3 & Cross-Subject & RGB+Radar & 47.48 & 35.74 \\
P3 & Cross-Subject & LiDAR+Radar & 45.83 & 35.97 \\

\bottomrule
\end{tabular}
}

\caption{
Bimodal fusion ablation on MM-Fi across the three activity protocols
and evaluation splits. Results are reported for RGB+LiDAR, RGB+Radar,
and LiDAR+Radar combinations. MPJPE and PA-MPJPE are reported in
millimeters; lower is better.
}
\label{tab:bimodal_ablation}

\end{table}

\subsection{Comparison with BeV fusion}
\label{sec:Bev}
\begin{table}[H]
\centering
\renewcommand{\arraystretch}{1.10}
\footnotesize
\setlength{\tabcolsep}{6pt}

\begin{tabular}{l l c}
\toprule
\textbf{Split} &
\textbf{Method} &
\textbf{MPJPE (mm)} $\downarrow$ \\
\midrule

Random
& BEV Baseline
& 67.03 \\

Random
& Our Fusion
& \textbf{43.21} \\

\midrule

Cross-Environment
& BEV Baseline
& 73.37 \\

Cross-Environment
& Our Fusion
& \textbf{67.03} \\

\midrule

Cross-Subject
& BEV Baseline
& 56.44 \\

Cross-Subject
& Our Fusion
& \textbf{45.83} \\

\bottomrule
\end{tabular}

\caption{
Comparison with the BEV-based fusion baseline \cite{ji2024vexkd} on MM-Fi Protocol 3 using the LiDAR+Radar bimodal setting across the random,
cross-environment, and cross-subject splits. MPJPE is reported in millimeters; lower is better.
}
\label{tab:bev_fusion_ablation}
\end{table}

\end{document}